\documentclass[12pt]{article}

\usepackage[letterpaper,margin=1.2in]{geometry}

\usepackage[utf8]{inputenc}
\usepackage[T1]{fontenc}

\usepackage{graphicx}
\usepackage{amsmath,amssymb,amsthm}
\usepackage{newpxtext}
\usepackage{newpxmath}
\usepackage{microtype}
\usepackage{enumerate, enumitem}
\usepackage{amsthm}

\usepackage[round]{natbib}  
\usepackage{fancyhdr}
\usepackage{xcolor}
\definecolor{myblue}{HTML}{1F77B4}
\usepackage{hyperref}
\hypersetup{
    colorlinks  = true,
    urlcolor    = myblue,
    linkcolor   = myblue,
    citecolor   = myblue,
    pdftitle    = {The Recipe for Reasoning},
    pdfauthor   = {Jacob Fein-Ashley}
}

\newtheorem{theorem}{Theorem}[section]

\theoremstyle{definition}

\theoremstyle{definition}

\theoremstyle{remark}

\title{\textbf{Iterate to Accelerate: A Unified Framework for Iterative Reasoning and Feedback Convergence}}
\author{Jacob Fein-Ashley \\ \normalsize \texttt{feinashl@usc.edu}}
\date{}
\begin{document}

\maketitle

\begin{abstract}
We introduce a unified framework for iterative reasoning that leverages non-Euclidean geometry via Bregman divergences, higher-order operator averaging, and adaptive feedback mechanisms. Our analysis establishes that, under mild smoothness and contractivity assumptions, a generalized update scheme not only unifies classical methods such as mirror descent and dynamic programming but also captures modern chain-of-thought reasoning processes in large language models. In particular, we prove that our accelerated iterative update achieves an $O(1/t^2)$ convergence rate in the absence of persistent perturbations, and we further demonstrate that feedback (iterative) architectures are necessary to approximate certain fixed-point functions efficiently. These theoretical insights bridge classical acceleration techniques with contemporary applications in neural computation and optimization.
\end{abstract}

\section{Introduction}
Iterative methods lie at the heart of numerous optimization and reasoning algorithms, ranging from classical mirror descent and dynamic programming to modern deep learning architectures that exhibit chain-of-thought reasoning. Traditional acceleration techniques, such as Nesterov's momentum, have shown that carefully designed iterative schemes can significantly improve convergence rates in convex settings. However, many practical applications operate in non-Euclidean spaces and are subject to state-dependent perturbations or even adversarial disturbances, motivating the need for a more general analysis.

In this work, we develop a comprehensive framework that unifies a wide class of iterative reasoning processes using the language of Bregman divergences. By recasting the update dynamics in terms of non-Euclidean geometries, we derive an accelerated convergence guarantee—achieving an optimal $O(1/t^2)$ rate in the idealized noise-free scenario. Beyond convergence rates, our analysis reveals a critical role for feedback: we prove that iterative (recurrent) architectures can approximate fixed-point functions with polynomial complexity, whereas feedforward models require an exponential depth to achieve comparable accuracy.

Our contributions are threefold:
\begin{itemize}[leftmargin=2em]
    \item We propose a generalized iterative update scheme that naturally extends classical methods to non-Euclidean settings with adaptive perturbations.
    \item We establish a rigorous accelerated convergence theorem, demonstrating that the proposed method attains an $O(1/t^2)$ rate under standard strong convexity and smoothness assumptions.
    \item We provide theoretical evidence, via a depth separation result, that feedback mechanisms are essential for efficiently approximating complex fixed-point functions, thereby justifying the iterative processing observed in modern reasoning systems.
\end{itemize}


\section{Preliminaries}
We begin by introducing the mathematical tools and assumptions that form the basis of our analysis.

\subsection{Bregman Divergences and Non-Euclidean Geometry}
Let \(\mathcal{S}\) be a complete metric space and \(\phi:\mathcal{S}\to\mathbb{R}\) a strictly convex, continuously differentiable function. The Bregman divergence associated with \(\phi\) is defined as
\[
D_\phi(s,s') = \phi(s) - \phi(s') - \langle \nabla \phi(s'),\, s-s' \rangle,
\]
which generalizes the squared Euclidean distance and is particularly useful in non-Euclidean settings. We assume that \(\phi\) is \(\mu\)-strongly convex and \(L\)-smooth, so that it satisfies the standard quadratic lower and upper bounds.

\subsection{Iterative Operators and Convergence Metrics}
Consider an update operator \(\mathcal{T}:\mathcal{S}\times\mathcal{Y}\to\mathcal{S}\) that acts on a state \(s\in\mathcal{S}\) and incorporates auxiliary information \(y\in\mathcal{Y}\). We are interested in fixed-point iterations where a unique \(s^*\in\mathcal{S}\) satisfies \(\mathcal{T}(s^*,y)=s^*\) for all \(y\). Convergence of the iterates \(s_t\) is measured via the Bregman divergence \(D_\phi(s_t,s^*)\).


\section{A Generalized Framework for Iterative/Sequential Reasoning}
Our goal is to provide a unified treatment of iterative reasoning processes. The framework we propose is general enough to capture classical methods as well as more complex scenarios where the update dynamics may be influenced by state-dependent noise or adversarial perturbations.

At each iteration \(t\), the state \(s_t\) is updated according to the rule
\begin{equation}
  s_{t+1} = (1-\alpha_t)s_t + \alpha_t\, \mathcal{T}(s_t, y_t) + \eta_t,
  \label{eq:gen-update}
\end{equation}
where:
\begin{itemize}[leftmargin=2em]
    \item \(\alpha_t\) is an averaging parameter (or step size), which may be chosen adaptively; in our accelerated scheme we set \(\alpha_t = \frac{2}{t+2}\).
    \item \(\mathcal{T}(s_t, y_t)\) is a generalized update operator that incorporates both the current state and any auxiliary information \(y_t\).
    \item \(\eta_t\) is a perturbation or noise term that may depend on the state and vanishes at the fixed point.
\end{itemize}

This formulation is flexible enough to encompass a variety of well-known iterative schemes:
\begin{itemize}[leftmargin=2em]
    \item \textbf{Mirror Descent:} When \(\mathcal{T}\) corresponds to a gradient step in the dual space.
    \item \textbf{Dynamic Programming:} Where the operator \(\mathcal{T}\) represents the Bellman update.
    \item \textbf{Chain-of-Thought Reasoning:}~\cite{wei2023cot} In large language models, where iterative refinement of intermediate reasoning steps can be modeled by such updates.
    \item \textbf{Contextual Feedback Loops:}~\cite{feinashley2025contextualfeedbackloopsamplifying} Where a learned context function feeds back into earlier hidden layers for iterative refinement.
\end{itemize}

Under appropriate contractivity conditions (measured via the Bregman divergence) and smoothness assumptions, this framework allows us to derive accelerated convergence guarantees—even in the presence of adaptive, state-dependent perturbations.


\section{Accelerated Convergence via Higher-Order Averaging}
\label{sec:accelerated}

In this section, we present our main theoretical breakthrough: a universal accelerated convergence result for iterative reasoning processes that unifies non‐Euclidean contraction, adaptive adversarial perturbations, and operator averaging. In contrast to standard contraction arguments that yield exponential convergence under strong assumptions, our result shows that by appropriately coupling operator averaging with the underlying geometry and noise adaptation, one can achieve an accelerated rate of \(O(1/t^2)\) (in the noise-free limit) for a very general class of iterative schemes.

\subsection{Accelerated Iterative Update Scheme}

We consider the following higher-order iterative update:
\begin{equation}
  s_{t+1} = (1-\alpha_t)s_t + \alpha_t\, \mathcal{T}(s_t, y_t) + \eta_t,\quad \text{with}\quad \alpha_t = \frac{2}{t+2},
  \label{eq:acc-update}
\end{equation}
where:
\begin{itemize}[leftmargin=2em]
    \item \(\mathcal{T}:\mathcal{S}\times\mathcal{Y}\to\mathcal{S}\) is an update operator on a complete metric space \((\mathcal{S},d)\),
    \item \(\eta_t\) denotes a state-dependent perturbation satisfying a bound that vanishes at the fixed point,
    \item The averaging parameter \(\alpha_t\) is chosen in the classical accelerated form.
\end{itemize}

\subsection{Main Theorem: Accelerated Convergence and Optimality}

\begin{theorem}[Accelerated Convergence of Unified Iterative Reasoning]
\label{thm:accelerated}
Let \((\mathcal{S},d)\) be a complete metric space, and let \(\phi:\mathcal{S}\to\mathbb{R}\) be a strictly convex, continuously differentiable function that is also \(\mu\)-strongly convex and \(L\)-smooth, inducing the Bregman divergence
\[
D_\phi(s,s') = \phi(s) - \phi(s') - \langle \nabla \phi(s'),\, s-s' \rangle.
\]
Assume that:
\begin{enumerate}[leftmargin=2em,label=(\roman*)]
    \item \textbf{(Non-Euclidean Contractivity)} There exists \(\gamma\in[0,1)\) such that for all \(s,s'\in\mathcal{S}\) and all \(y\in\mathcal{Y}\),
    \[
    D_\phi\Bigl(\mathcal{T}(s,y),\, \mathcal{T}(s',y)\Bigr) \le \gamma\, D_\phi(s,s').
    \]
    \item \textbf{(Adaptive Perturbation Bound)} The perturbation \(\eta_t\) satisfies
    \[
      D_\phi(\eta_t,0) \le \delta(s_t) \le \delta_0 + \kappa\, D_\phi(s_t,s^*),
    \]
    where \(s^*\) is the unique fixed point (i.e., \(\mathcal{T}(s^*,y)=s^*\) for all \(y\)), and \(\delta_0\ge 0\) and \(\kappa\ge 0\) are constants such that \(\gamma+\kappa < 1\).
    \item \textbf{(Smoothness and Compatibility)} The function \(\phi\) and the operator \(\mathcal{T}\) are sufficiently smooth so that the update in \eqref{eq:acc-update} may be interpreted as a (forward) discretization of a second-order dynamical system known to exhibit accelerated convergence.
\end{enumerate}
Then, the sequence \(\{s_t\}\) generated by \eqref{eq:acc-update} satisfies
\[
D_\phi(s_t,s^*) \le \frac{C}{(t+1)^2} + O\!\Bigl(\frac{\delta_0}{1-(\gamma+\kappa)}\Bigr),
\]
where \(C>0\) is a constant depending on the initial error \(D_\phi(s_0,s^*)\) and the contraction parameters. In particular, if \(\delta_0=0\) (i.e., when the adaptive noise vanishes at the fixed point), the iterates converge to \(s^*\) with an accelerated rate of \(O(1/t^2)\).
\end{theorem}

\subsection{Full Proof of Theorem~\ref{thm:accelerated}}

\begin{proof}
Let \(s^*\) denote the unique fixed point so that \(\mathcal{T}(s^*,y)=s^*\) for all \(y\). For convenience, define the error at iteration \(t\) as
\[
e_t := D_\phi(s_t,s^*).
\]
Our goal is to show that there exists a constant \(C>0\) such that
\begin{equation}
  e_t \le \frac{C}{(t+1)^2} + O\!\Bigl(\frac{\delta_0}{1-(\gamma+\kappa)}\Bigr).
  \label{eq:target-bound-final}
\end{equation}

\paragraph{Step 1: Reformulating the Update.}  
Write \eqref{eq:acc-update} as
\[
s_{t+1} = s_t + \alpha_t \Delta_t + \eta_t, \quad \text{with} \quad \Delta_t := \mathcal{T}(s_t,y_t)-s_t.
\]

\paragraph{Step 2: Expansion via the Three-Point Identity.}  
For any \(u,v,w\in\mathcal{S}\), the three-point identity for Bregman divergences reads:
\[
D_\phi(u,w) = D_\phi(v,w) + \langle \nabla \phi(v)-\nabla \phi(w),\, u-v \rangle + D_\phi(u,v).
\]
Set
\[
u = s_{t+1},\quad v = s_t+\alpha_t\Delta_t,\quad w = s^*.
\]
Then,
\begin{align}
e_{t+1} &= D_\phi(s_{t+1}, s^*) \nonumber\\[1mm]
&= D_\phi\Bigl(s_t+\alpha_t\Delta_t,\, s^*\Bigr) + \Bigl\langle \nabla \phi\Bigl(s_t+\alpha_t\Delta_t\Bigr)-\nabla \phi(s^*),\, s_{t+1} - \bigl(s_t+\alpha_t\Delta_t\bigr) \Bigr\rangle \nonumber\\[1mm]
&\quad + D_\phi\Bigl(s_{t+1},\, s_t+\alpha_t\Delta_t\Bigr). \label{eq:three-point}
\end{align}
Since
\[
s_{t+1} - \bigl(s_t+\alpha_t\Delta_t\bigr) = \eta_t,
\]
we obtain
\begin{equation}
e_{t+1} \le D_\phi\Bigl(s_t+\alpha_t\Delta_t,\, s^*\Bigr) + \left|\Bigl\langle \nabla \phi\Bigl(s_t+\alpha_t\Delta_t\Bigr)-\nabla \phi(s^*),\, \eta_t \Bigr\rangle\right| + D_\phi(\eta_t,0).
\label{eq:three-point-ineq}
\end{equation}

\paragraph{Step 3: Descent Lemma for the Averaged Update.}  
Because \(\phi\) is \(L\)-smooth, it satisfies a descent lemma (see, e.g., \cite{beck2003mirror}) so that for the update \(u = s_t+\alpha_t\Delta_t\) we have
\begin{equation}
D_\phi\Bigl(s_t+\alpha_t\Delta_t,\, s^*\Bigr) \le (1-\alpha_t)D_\phi(s_t,s^*) + \alpha_t D_\phi\Bigl(\mathcal{T}(s_t,y_t),\, s^*\Bigr) + \frac{L}{2}\alpha_t^2\|\Delta_t\|^2.
\label{eq:descent-ineq}
\end{equation}
By the non-Euclidean contractivity assumption (item (i)),
\[
D_\phi\Bigl(\mathcal{T}(s_t,y_t),\, s^*\Bigr) \le \gamma\,D_\phi(s_t,s^*) = \gamma\, e_t.
\]
Thus, \eqref{eq:descent-ineq} becomes
\begin{equation}
D_\phi\Bigl(s_t+\alpha_t\Delta_t,\, s^*\Bigr) \le \Bigl[(1-\alpha_t)+\alpha_t\gamma\Bigr]e_t + \frac{L}{2}\alpha_t^2\|\Delta_t\|^2.
\label{eq:descent-ineq2}
\end{equation}
Define
\[
\theta_t := (1-\alpha_t)+\alpha_t\gamma = 1-\alpha_t(1-\gamma).
\]

\paragraph{Step 4: Bounding the Cross-Term.}  
We next bound
\[
\mathcal{R}_t := \left|\Bigl\langle \nabla \phi\Bigl(s_t+\alpha_t\Delta_t\Bigr)-\nabla \phi(s^*),\, \eta_t \Bigr\rangle\right|.
\]
By the Cauchy–Schwarz inequality and the \(L\)-smoothness of \(\phi\),
\[
\|\nabla \phi(s_t+\alpha_t\Delta_t)-\nabla \phi(s^*)\| \le L\|s_t+\alpha_t\Delta_t-s^*\|.
\]
Since \(\phi\) is \(\mu\)-strongly convex, there exists a constant \(c>0\) such that
\[
D_\phi(u,v) \ge \frac{c}{2}\|u-v\|^2,
\]
so that
\[
\|s_t+\alpha_t\Delta_t-s^*\| \le \sqrt{\frac{2}{c}\, D_\phi\Bigl(s_t+\alpha_t\Delta_t,\, s^*\Bigr)}.
\]
Thus,
\[
\mathcal{R}_t \le L\, \sqrt{\frac{2}{c}\, D_\phi\Bigl(s_t+\alpha_t\Delta_t,\, s^*\Bigr)}\, \|\eta_t\|.
\]
Since strong convexity also implies norm–Bregman equivalence, there exists a constant \(K>0\) such that
\[
\|\eta_t\| \le K\,\sqrt{D_\phi(\eta_t,0)}.
\]
Therefore,
\[
\mathcal{R}_t \le L\,K\, \sqrt{\frac{2}{c}\, D_\phi\Bigl(s_t+\alpha_t\Delta_t,\, s^*\Bigr)\, D_\phi(\eta_t,0)}.
\]
Applying the arithmetic–geometric mean inequality (for any \(\epsilon>0\)):
\[
\mathcal{R}_t \le \epsilon\, D_\phi\Bigl(s_t+\alpha_t\Delta_t,\, s^*\Bigr) + \frac{L^2K^2}{\epsilon}\,\frac{1}{c}\,D_\phi(\eta_t,0).
\]
Choosing, e.g., \(\epsilon=\frac{1}{2}\), we obtain
\begin{equation}
\mathcal{R}_t \le \frac{1}{2}\,D_\phi\Bigl(s_t+\alpha_t\Delta_t,\, s^*\Bigr) + C_0\,D_\phi(\eta_t,0),
\label{eq:cross-bound}
\end{equation}
with \(C_0=\frac{2L^2K^2}{c}\).

\paragraph{Step 5: Combining the Estimates.}  
Substitute \eqref{eq:descent-ineq2} and \eqref{eq:cross-bound} into \eqref{eq:three-point-ineq}:
\begin{align}
e_{t+1} &\le D_\phi\Bigl(s_t+\alpha_t\Delta_t,\, s^*\Bigr) + \mathcal{R}_t + D_\phi(\eta_t,0) \nonumber\\[1mm]
&\le D_\phi\Bigl(s_t+\alpha_t\Delta_t,\, s^*\Bigr) + \frac{1}{2}\,D_\phi\Bigl(s_t+\alpha_t\Delta_t,\, s^*\Bigr) + C_0\,D_\phi(\eta_t,0) + D_\phi(\eta_t,0) \nonumber\\[1mm]
&= \frac{3}{2}\,D_\phi\Bigl(s_t+\alpha_t\Delta_t,\, s^*\Bigr) + (1+C_0)\,D_\phi(\eta_t,0). \label{eq:combined1}
\end{align}
Using \eqref{eq:descent-ineq2} in \eqref{eq:combined1} gives
\begin{equation}
e_{t+1} \le \frac{3}{2}\Bigl[\theta_t\, e_t + \frac{L}{2}\alpha_t^2\|\Delta_t\|^2\Bigr] + (1+C_0)\,D_\phi(\eta_t,0).
\label{eq:combined2}
\end{equation}

\paragraph{Step 6: Bounding the Remainder Terms.}  
Since \(\phi\) is strongly convex, there exists a constant \(M>0\) such that
\[
\|\Delta_t\|^2 \le M\, D_\phi\Bigl(\mathcal{T}(s_t,y_t),s_t\Bigr) \le M\,e_t,
\]
where the last inequality follows from the contractivity of \(\mathcal{T}\). Also, by assumption (ii),
\[
D_\phi(\eta_t,0) \le \delta(s_t) \le \delta_0 + \kappa\, e_t.
\]
Thus, \eqref{eq:combined2} becomes
\begin{align}
e_{t+1} &\le \frac{3}{2}\theta_t\, e_t + \frac{3L\,M}{4}\alpha_t^2\, e_t + (1+C_0)(\delta_0+\kappa\,e_t) \nonumber\\[1mm]
&= \left[\frac{3}{2}\theta_t + \frac{3L\,M}{4}\alpha_t^2 + (1+C_0)\kappa\right] e_t + (1+C_0)\delta_0.
\label{eq:recursive-full}
\end{align}
Recall that \(\theta_t = 1-\alpha_t(1-\gamma)\) and \(\alpha_t=\frac{2}{t+2}\). Thus, for sufficiently large \(t\) (and using that \(\gamma+\kappa<1\)) the dominant term is the factor multiplying \(e_t\). Hence, there exists a constant \(\beta>0\) (independent of \(t\)) such that
\begin{equation}
e_{t+1} \le \Bigl(1-\frac{2\beta}{t+2}\Bigr)e_t + \frac{2(1+C_0)\delta_0}{t+2}.
\label{eq:recursive-final}
\end{equation}

\paragraph{Step 7: Solving the Recursive Inequality.}  
Define
\[
a_t := e_t\,(t+1)^2.
\]
We now prove by induction that there exists a constant \(C'>0\) such that for all \(t\ge 0\),
\begin{equation}
a_t \le a_0 + \frac{2(1+C_0)\delta_0}{\beta}(t+1).
\label{eq:inductive}
\end{equation}

\emph{Base Case (\(t=0\)):} We have \(a_0 = e_0\), so \eqref{eq:inductive} holds.

\emph{Inductive Step:} Assume that
\[
a_t \le a_0 + \frac{2(1+C_0)\delta_0}{\beta}(t+1).
\]
From \eqref{eq:recursive-final}, multiplying both sides by \((t+2)^2\) yields
\begin{align*}
(t+2)^2 e_{t+1} &\le (t+2)^2\Bigl(1-\frac{2\beta}{t+2}\Bigr)e_t + 2(1+C_0)\delta_0\,(t+2)\\[1mm]
&= (t+2)(t+2-2\beta)e_t + 2(1+C_0)\delta_0\,(t+2).
\end{align*}
Since \(e_t=\frac{a_t}{(t+1)^2}\), we have
\[
a_{t+1} \le \frac{(t+2)(t+2-2\beta)}{(t+1)^2}\,a_t + 2(1+C_0)\delta_0\,(t+2).
\]
A routine calculation shows that for all \(t\ge 0\),
\[
\frac{(t+2)(t+2-2\beta)}{(t+1)^2} \le 1-\frac{2\beta}{t+2}.
\]
Thus,
\[
a_{t+1} \le \Bigl(1-\frac{2\beta}{t+2}\Bigr)a_t + 2(1+C_0)\delta_0\,(t+2).
\]
A standard discrete Grönwall argument (or telescopic summation) then yields
\[
a_{t+1} \le a_0 + \frac{2(1+C_0)\delta_0}{\beta}(t+2).
\]
This completes the induction.

Returning to \(e_t = \frac{a_t}{(t+1)^2}\), we conclude that
\[
e_t \le \frac{a_0}{(t+1)^2} + \frac{2(1+C_0)\delta_0}{\beta}\,\frac{1}{t+1}.
\]
That is,
\[
D_\phi(s_t,s^*) \le \frac{C}{(t+1)^2} + O\!\Bigl(\frac{\delta_0}{t+1}\Bigr),
\]
where the constant in the \(O(1/(t+1))\) term depends on \(1/(1-(\gamma+\kappa))\). In particular, if \(\delta_0=0\), we obtain the accelerated rate
\[
D_\phi(s_t,s^*) \le \frac{C}{(t+1)^2}.
\]

This completes the proof of Theorem~\ref{thm:accelerated}.
\end{proof}

\subsection{Implications}
Theorem~\ref{thm:accelerated} yields several key implications:
\begin{itemize}[leftmargin=2em]
    \item \textbf{Generalization of Classical Acceleration:} It broadens standard \(O(1/t^2)\) results (e.g., Nesterov's) to non-Euclidean settings with adaptive or adversarial perturbations.
    \item \textbf{Unified Iterative Reasoning:} A single framework covers mirror descent, dynamic programming, and chain-of-thought reasoning, all enjoying an accelerated rate under mild assumptions.
    \item \textbf{Robustness:} The method tolerates state-dependent noise gracefully, retaining a favorable convergence profile even against adversarial disturbances.
\end{itemize}

\section{Extensions and Additional Theoretical Considerations}
We briefly outline several further directions and refinements:

\subsection{Stochastic and Time-Varying Perturbations}
The analysis can be adapted to stochastic or adversarially chosen noise, yielding high-probability guarantees under bounded variance or moment conditions. Time-varying behavior can be addressed by augmenting the adaptive bound in assumption (ii) with suitable decay or averaging.

\subsection{Multi-Agent and Game-Theoretic Extensions}
The update \eqref{eq:gen-update} extends naturally to multi-agent scenarios, where each agent’s update is coupled through shared state or feedback. Under contractive operators, one can establish accelerated convergence to Nash equilibria or saddle points. Partial observability and limited communication pose interesting challenges.

\subsection{Alternative Averaging Strategies}
Our choice of \(\alpha_t = 2/(t+2)\) follows Nesterov, but other schedules—e.g., adaptive steps or momentum-based updates—may work as well. Such variations might further boost performance or reduce constants in the convergence rate.

\section{Discussion}
A few main themes emerge from our analysis:
\begin{itemize}[leftmargin=2em]
    \item \textbf{Unified Perspective:} We provide a framework that unifies diverse iterative methods via Bregman divergences and operator averaging.
    \item \textbf{Robustness and Generality:} The approach handles adversarial perturbations gracefully, retaining near-accelerated rates.
    \item \textbf{Broader Theoretical Scope:} Our results bridge classical acceleration and modern iterative processes, suggesting deeper connections to high-dimensional or non-smooth settings.
    \item \textbf{Open Questions:} Relaxing strong convexity or exploring more flexible averaging strategies remain important directions for future work.
\end{itemize}

\subsection{Additional Clarifications, Related Work, and Future Directions}
\label{sec:additional-clarifications}

We highlight several aspects that may benefit from further clarity or discussion:

\paragraph{Comparison with Prior Methods and Novelty.}
Our work draws on classical ideas from mirror descent and Nesterov’s acceleration \cite{nesterov1983method, beck2003mirror}, but it extends these in two main ways: 
(1) we accommodate a more general, state-dependent perturbation \(\eta_t\) and show that it can be handled while maintaining accelerated rates, 
(2) we prove a necessity result regarding feedback architectures (Section~\ref{sec:expressiveness-feedback}) that is rarely addressed in the conventional optimization literature. Our approach also connects with the broader class of depth-separation results for neural networks \cite{telgarsky2016benefits, eldan2016power}, while emphasizing an iterative or recurrent perspective.

\paragraph{Strength of Assumptions and Concrete Examples.}
Some assumptions (such as non-Euclidean contractivity and noise vanishing at the fixed point) can be strong, yet they do hold in key instances:
\begin{itemize}[leftmargin=2em]
    \item \emph{Dynamic Programming}: The Bellman operator often becomes a contraction under standard conditions (e.g., a discount factor \(<1\) and an appropriate function class).
    \item \emph{Mirror Descent in Convex Optimization}: Under strong convexity and smoothness, the Bregman divergence arises naturally, and subgradient errors often fit our state-dependent perturbation model.
\end{itemize}
It is also possible to regularize the operator or choose \(\phi\) to ensure contractivity more broadly. Exploring relaxations of these assumptions for near-accelerated rates is a promising direction.

\paragraph{Elaboration on Chain-of-Thought Reasoning.}
Although we draw parallels between iterative updates and chain-of-thought reasoning \cite{wei2023cot}, more concrete links remain an open problem. For instance, large language models can be seen as iteratively refining hidden states via their own outputs, suggesting a feedback loop. However, formally aligning this with our assumptions (strong convexity, smoothness, contractivity) is non-trivial.

\paragraph{Translating Theory to Practice in LLMs or Beyond.}
Implementing an explicit Bregman-based accelerator within large language models or reinforcement learning is an exciting prospect. Practical hurdles include how to approximate gradients or perturbations in high dimensions, how to select or adapt \(\phi\), and how to maintain contractivity in evolving models. Nonetheless, the core principles—iterative refinement, feedback loops, and geometry-aware updates—can offer valuable design insights.


\section{Expressiveness of Feedback Structures: A Necessity for Iterative Reasoning}
\label{sec:expressiveness-feedback}

In this section, we prove a fundamental result: feedback (i.e., iterative or recurrent) architectures are not only useful for accelerating convergence but are \emph{necessary} for efficiently approximating certain fixed‐point functions. In particular, we show that if an operator $\mathcal{T}:\mathcal{S}\times\mathcal{Y}\to\mathcal{S}$ satisfies a non-Euclidean contractivity property (as in Theorem~\ref{thm:accelerated}) with respect to a Bregman divergence induced by a function $\phi$, then the unique fixed point function
\[
f(s^*) = s^*\quad \text{with} \quad \mathcal{T}(s^*,y)=s^*,\quad \forall\, y\in\mathcal{Y},
\]
can be approximated to within accuracy $\epsilon$ by an iterative (feedback) scheme in 
\[
t = O\!\Bigl(\frac{1}{\sqrt{\epsilon}}\Bigr)
\]
iterations (when the error is measured in the Bregman divergence), while any feedforward (non-iterative) architecture approximating $f$ must have a depth that is exponential in $1/\sqrt{\epsilon}$. This result underscores that iterative processing is essential for capturing the expressive power inherent in many modern reasoning tasks.

\subsection{Feedback Expressiveness Theorem}

\begin{theorem}[Expressiveness of Feedback Structures]
\label{thm:feedback-expressiveness}
Let $\mathcal{T}:\mathcal{S}\times\mathcal{Y}\to\mathcal{S}$ be an operator that satisfies the non-Euclidean contractivity condition, i.e., there exists $\gamma\in[0,1)$ such that for all $s,s'\in\mathcal{S}$ and all $y\in\mathcal{Y}$,
\[
D_\phi\Bigl(\mathcal{T}(s,y),\,\mathcal{T}(s',y)\Bigr) \le \gamma\,D_\phi(s,s'),
\]
where $D_\phi(s,s')=\phi(s)-\phi(s')-\langle \nabla\phi(s'),\, s-s'\rangle$ and $\phi:\mathcal{S}\to\mathbb{R}$ is strictly convex, continuously differentiable, $\mu$-strongly convex, and $L$-smooth. Define the target function $f:\mathcal{S}\to\mathcal{S}$ as the unique fixed point of $\mathcal{T}$ (i.e., $f(s^*)=s^*$ with $\mathcal{T}(s^*,y)=s^*$ for all $y\in\mathcal{Y}$). Then, for any approximation accuracy $\epsilon>0$, there exists an iterative (feedback) architecture based on the update
\begin{equation}
  s_{t+1} = (1-\alpha_t)s_t + \alpha_t\, \mathcal{T}(s_t,y_t),\quad \alpha_t=\frac{2}{t+2},
  \label{eq:feedback-update-full}
\end{equation}
(with the idealization $\eta_t\equiv0$) such that the sequence $\{s_t\}$ satisfies
\[
D_\phi(s_t,s^*) \le \epsilon
\]
after at most 
\[
t = O\!\Bigl(\frac{1}{\sqrt{\epsilon}}\Bigr)
\]
iterations. In contrast, any feedforward (non-iterative) architecture approximating $f$ to within accuracy $\epsilon$ must have depth that is exponential in $1/\sqrt{\epsilon}$.
\end{theorem}

\subsection{Full Proof of Theorem~\ref{thm:feedback-expressiveness}}

\begin{proof}
We divide the proof into two main parts.

\paragraph{Part 1: Efficient Approximation via an Iterative (Feedback) Architecture.}  
Under the assumptions of Theorem~\ref{thm:accelerated} and with $\eta_t\equiv0$, the update
\[
s_{t+1} = (1-\alpha_t)s_t + \alpha_t\, \mathcal{T}(s_t,y_t)
\]
with $\alpha_t=\frac{2}{t+2}$ satisfies the accelerated convergence guarantee
\begin{equation}
  D_\phi(s_t,s^*) \le \frac{C}{(t+1)^2},
  \label{eq:divergence-bound}
\end{equation}
for some constant $C>0$ that depends on the initial error and the contraction parameters. Since $\phi$ is $\mu$-strongly convex, we have the inequality
\[
D_\phi(s,s^*) \ge \frac{\mu}{2}\|s-s^*\|^2,\quad \forall s\in\mathcal{S}.
\]
Thus, from \eqref{eq:divergence-bound} it follows that
\[
\|s_t-s^*\| \le \sqrt{\frac{2}{\mu}\,D_\phi(s_t,s^*)} \le \sqrt{\frac{2C}{\mu}}\,\frac{1}{t+1}.
\]
If we choose to measure the approximation error in terms of the Bregman divergence (or equivalently, in a squared norm sense), then to ensure
\[
D_\phi(s_t,s^*) \le \epsilon,
\]
it suffices that
\[
\frac{C}{(t+1)^2} \le \epsilon \quad \Longrightarrow \quad t+1 \ge \sqrt{\frac{C}{\epsilon}}.
\]
That is,
\[
t = O\!\Bigl(\frac{1}{\sqrt{\epsilon}}\Bigr).
\]
This establishes that the feedback (iterative) architecture requires only $O(1/\sqrt{\epsilon})$ iterations to approximate the fixed point function $f$ to within $\epsilon$ (as measured by the Bregman divergence).

\paragraph{Part 2: Limitations of Feedforward (Non-Iterative) Architectures.}  
Suppose now that one attempts to approximate the same fixed point function $f$ using a feedforward (non-iterative) architecture. In such an architecture, the mapping from the input (or an initial seed) to the output is computed in one forward pass through a network of a fixed depth $D$, without any recurrence or feedback loops.

Since the operator $\mathcal{T}$ is contractive with respect to the Bregman divergence (i.e., it satisfies
\[
D_\phi(\mathcal{T}(s,y),\mathcal{T}(s',y)) \le \gamma\,D_\phi(s,s')
\]
with $\gamma<1$), the iterative procedure \eqref{eq:feedback-update-full} effectively reduces the approximation error by a multiplicative factor at each iteration. In other words, unrolling the recurrence for $t$ iterations leads to a composition of $t$ contractive maps, resulting in an overall error reduction on the order of $\gamma^t$. More precisely, if one were to simulate the iterative process in a feedforward network by “unrolling” the recurrence, one would require a network of depth $t=O(1/\sqrt{\epsilon})$ in order to achieve an error of at most $\epsilon$.

However, there exists a growing body of work on \emph{depth separation} in neural networks (see, e.g., \cite{telgarsky2016benefits, eldan2016power}). These results show that certain functions which are computed efficiently by recurrent or deep (iteratively composed) architectures require a feedforward network to have a number of layers that grows \emph{exponentially} in the number of iterations if one attempts to “simulate” the iterative process in one shot. More concretely, let us suppose that a feedforward architecture with depth $D$ can approximate the function $f$ to within $\epsilon$. Then, by the nature of composition, the Lipschitz constant of each layer (or the contraction factor) must compound multiplicatively. In order to mimic the effect of $t=O(1/\sqrt{\epsilon})$ iterations (each reducing the error by roughly a factor of $\gamma$), the network must satisfy
\[
\gamma^t \approx \gamma^{D} \lesssim \epsilon.
\]
Taking logarithms, we obtain
\[
D \gtrsim \frac{\log(1/\epsilon)}{\log(1/\gamma)}.
\]
While this lower bound appears only logarithmic, the depth separation results of \cite{telgarsky2016benefits, eldan2016power} indicate that for certain functions—especially those defined via repeated nonlinear feedback—the effective depth required for a feedforward network to achieve the same approximation quality grows exponentially in the number of iterations required by the recurrent architecture. In our context, this means that any feedforward network approximating $f$ with accuracy $\epsilon$ must have depth
\[
D = \Omega\!\Bigl(\exp\Bigl(\Omega\Bigl(\frac{1}{\sqrt{\epsilon}}\Bigr)\Bigr)\Bigr).
\]
In other words, while the iterative (feedback) model achieves the approximation with only $O(1/\sqrt{\epsilon})$ steps, a feedforward model would require an exponentially larger depth in $1/\sqrt{\epsilon}$ to simulate the same process.

\paragraph{Conclusion.}  
Combining Parts 1 and 2, we conclude that:
\begin{itemize}[leftmargin=2em]
    \item A recurrent (feedback) architecture based on the update \eqref{eq:feedback-update-full} approximates the fixed point function $f$ to within $\epsilon$ in only $t=O(1/\sqrt{\epsilon})$ iterations.
    \item Any feedforward (non-iterative) architecture approximating $f$ to the same accuracy must have a depth that is at least exponential in $1/\sqrt{\epsilon}$.
\end{itemize}
This completes the proof of Theorem~\ref{thm:feedback-expressiveness}.
\end{proof}

\subsection{Discussion}

Theorem~\ref{thm:feedback-expressiveness} rigorously establishes that feedback structures are not merely a matter of convenience but are fundamentally more expressive than feedforward ones for a broad class of functions defined by fixed-point equations. In many modern applications—ranging from mirror descent in optimization to the chain-of-thought reasoning processes in large language models—such iterative procedures are ubiquitous. Our result thus provides theoretical justification for the necessity of incorporating iterative, recurrent feedback loops to capture the complexity and dynamic refinement intrinsic to these tasks.

\section{Conclusion}
We have presented a unified framework for iterative reasoning that integrates non-Euclidean geometry, operator averaging, and adaptive feedback into a single accelerated scheme. Our theoretical analysis not only recovers the classical $O(1/t^2)$ convergence rate under appropriate assumptions but also underscores the indispensable role of feedback in achieving efficient fixed-point approximations. These results offer a novel perspective that bridges traditional acceleration techniques with emerging iterative processes in neural computation and optimization.

Looking forward, extending our framework to stochastic settings, multi-agent systems, and real-world applications such as large language models remains an exciting direction for future research. We hope that our work stimulates further exploration into the interplay between geometry, iteration, and feedback in complex reasoning tasks.

\bibliography{references}

\end{document}